\documentclass[journal,comsoc]{IEEEtran}
\usepackage[T1]{fontenc}
\usepackage{graphicx}
\usepackage{caption}
\graphicspath{{./Figures/}}
\ifCLASSINFOpdf
\else
\fi
\usepackage{amsmath}
\usepackage[cmintegrals]{newtxmath}
\usepackage{comment}
\providecommand{\keywords}[1]
{
  \small	
  \textbf{\textit{Keywords---}} #1
}

\begin{document}

\title{Anomaly Detection in Residential Video Surveillance on Edge Devices in IoT Framework}

\author{Mayur R. Parate,
        Kishor M. Bhurchandi,~\IEEEmembership{Member~IEEE,}
        and~Ashwin G. Kothari,~\IEEEmembership{Senior Member,~IEEE}
        
\thanks{Mayur R. Parate is with the Department
of Electronics and Communication Engineering, Indian Institute of Information Technology, Nagpur, India. e-mail:mparate@iitn.ac.in}
\thanks{Kishor M. Bhurchandi and Ashwin G. Kothari are with Visvesvaraya National Institute of Technology, Nagpur, India.}
}

\maketitle

\begin{abstract}
 Intelligent resident surveillance is one of the most essential smart community services. The increasing demand for security needs surveillance systems to be able to detect anomalies in surveillance scenes. Employing high-capacity computational devices for intelligent surveillance in residential societies is costly and not feasible. Therefore, we propose anomaly detection for intelligent surveillance using CPU-only edge devices. A modular framework to capture object-level inferences and tracking is developed. To cope with partial occlusions, posture deformations, and complex scenes, we employed feature encoding and trajectory association governed by two metrices complementing to each other. The elements of an anomaly detection framework are optimized to run on CPU-only edge devices with sufficient frames per second (FPS). The experimental results indicate the proposed method is feasible and achieves satisfactory results in real-life scenarios.
\end{abstract}


\section{Introduction}
\IEEEPARstart{T}{he} recent concept of Smart Cities influences the town planners, investigators and researchers to strengthen the security and safety of a resident. Increasing demand for residential security needs an infrastructure of surveillance cameras deployed for video analysis. A leading challenge in analyzing surveillance videos is the detection of abnormal events or anomalies which require exhaustive human efforts. 
Most of these surveillance applications create massive contextual data that require significant storage and computing resources. An Internet-of-Things (IoT) framework provides excellent flexibility and scalability to handle increasing surveillance needs. Embedding smart surveillance in IoT framework needs computer vision tasks such as human detection, classification, and tracking to be performed essentially at the perception or sensor layer of an IoT architecture to ensure response time, accuracy, and energy efficiency. In safety-critical applications such as anomaly detection, response time is an important parameter. The visual tasks at the sensor layer of an IoT architecture depend on the ability to capture, process visual data, and making decisions in real-time. Thus, it solves the application problem using only a single available frame without storing the intermediate frames for processing. This need is not fulfilled by cloud computing, although high accuracy can be achieved for visual tasks. The need of extensive bandwidth and inevitable network latency will prevent real-time response for targeting moving objects \cite{kumaran2019anomaly}. Some approaches have been proposed for combining the edge and cloud to reduce network traffic and latencies \cite{ghosh2020edge}, however, due to the data-intensive nature of computer vision tasks, the network latencies are inevitable. 

Convolutional Neural Networks (CCN) are becoming popular as they provide near-human accuracy for object detection and tracking. Convolutional Siamese Network, a type of CNN is used in modern object detection and tracking, establishing benchmark \cite{8649753}, \cite{nawaratne2019spatiotemporal}, \cite{DBLP:journals/corr/abs-1810-04108}. However, to achieve the required accuracy, CNN-based object detection and tracking depend on the high processing power of underlying hardware/computers. They involve a huge quantity of computations making them unsuitable for edge devices. 
Employing edge computing for surveillance is considered as the answer to these shortcomings \cite{8422970}, \cite{patrikar2021anomaly}. Consequently, its integration with the IoT framework possesses the following advantages: real-time response, reduced network workload, lower power consumption, and higher data security and privacy. Despite the promising benefits of edge computing, one of the critical challenges is how to efficiently process the data-intensive computer vision tasks in real-time on a resource-constrained device. Various smart video surveillance approaches for object detection and tracking propose to use Artificial Intelligence (AI), Machine Learning (ML), and Deep Learning (DL) algorithms and they usually have huge computational requirements. How to migrate these data-intensive and high-computing tasks to the edge devices at IoT nodes are still significant challenge.
The research contributions of this work are as follows:
\begin{enumerate}
    \item A lightweight framework for anomaly detection in residential video surveillance is developed for CPU-only edge devices to achieve real-time performance.
    \item We emoloy spatio-temporal features for trajectory association based on two metrics to cope with partial occlusions, posture deformations, and complex scenes.
    \item We optimized framework elements to run on CPU-only edge device and effect of configuration parameters on the overall accuracy and efficiency of the anomaly detection system is analyzed.
    \item We integrated the surveillance system into an IoT framework for real-time automated alerts and analyzed the performance of anomaly detector on real-world residential surveillance video streams for CPU-only edge devices.
\end{enumerate}

The rest of the paper is organized as follows: in section II we present the learning-based approaches for anomaly detection and then the attempts made for anomaly detection at the Edge/terminal devices are discussed. Section III presents the methodology and framework for the proposed anomaly detection. Experimental analysis for different configurations of parameters is presented in Section IV. Results and benchmarking on standard datasets are presented in Section V. Finally, Section VI concludes this paper.

\section{Related Work}
Currently, most of the video surveillance systems function as an archive of footage and need huge storage and processing in the background. Very few are real-time smart surveillance systems that perform object detection and tracking tasks with the backbone of networking servers and Cloud operations. It has been recognized that heavy communication overhead observed in the cloud is not tolerable in many delay-sensitive, mission-critical tasks such as anomaly detection \cite{7979979}. The terminal processing at edge devices for automated surveillance is considered to be the alternative for cloud and network-based processing \cite{8422970}, especially when response time is a relatively important parameter. Over the decade, some approaches have been crafted for automated surveillance/object tracking using edge devices. However, very few of them talk about the detection of an anomaly in video surveillance.\par

\begin{figure}[t]
\includegraphics[width=0.8\linewidth]{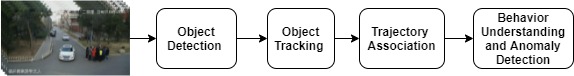}
\centering
\caption{General flow of Anomaly Detection}
\end{figure}

Though the notion of anomaly is not the same in all application contexts \cite{6544585}, \cite{sodemann2012review}, \cite{loce2013computer} but it follows a general framework comprising mainly; object detection, tracking, trajectory association, behavior understanding, and abnormal event/anomaly detection as shown in Fig. 1.\par
Object detection and tracking solutions range from traditional computer vision algorithms to more recent applications of learning-based approaches. Traditional algorithms; Scale Invariance Feature Transformation (SIFT) performs human detection using distinctive invariant features that are used to reliably match different views of an object or scene \cite{lowe2004distinctive}. Histograms of Oriented Gradient (HOG) descriptors significantly outperform existing feature sets for human detection \cite{1467360}. Further, HOG+SVM algorithm \cite{pang2011efficient} shows better performance in human detection. To detect crowd anomalies, histogram of magnitude and momentum \cite{bansod2020crowd}, Spatio-temporal motion pattern models with Kernelized SVM \cite{priyadharsini2021kernel} are utilized. Simone et al. modeled the bio-inspired interaction to understand  behaviour and anomalies in the crowd \cite{chiappino2015bio}. Whereas, the potential of modern learning-based strategies has led computer vision to a new extent during the last few years \cite{nature521}. Deep learning methods and Convolutional Neural Networks (CNN) along with recent computational advancement has opened a new way to handle computer vision tasks such as object tracking \cite{DBLP:journals/corr/GundogduA17}. Although the training phase usually demands high computation capability, the benefits of CNN can be exploited using computationally constrained hardware resources once the CNN has been trained. Few approaches have proposed the hardware-efficient neural networks to deliver state-of-the-art performance on embedded systems \cite{Zhang2020SkyNetAH}, \cite{8945145}, \cite{7533003}, \cite{howard2017mobilenets}.
Recently, learning-based approaches specifically designed for anomaly detection are observed in the literature \cite{8649753}, \cite{nawaratne2019spatiotemporal}, \cite{9116533}. \cite{wei2018unsupervised} detected anomalies in traffic surveillance, utilizing background modeling followed by Faster RCNN to detect vehicles in the extracted background and decide new anomalies under certain conditions. Following an unsupervised approach, \cite{bhakat2019anomaly} used an autoencoder model trained to minimize the reconstruction error between the input and the generated output. Further, the scalability issues due to the different-sized anomaly and their occurrence in a short time in pedestrian pathways are resolved by Region-based Scalable Convolution Neural Network (RS-CNN) \cite{murugan2019region}. An AnomalyNet in \cite{8649753} detects anomaly by the synergic association of feature learning, sparse representation using sparse long short term memory (SLSTM). Despite the rapid development computer vision applications using deep learning methods and CNNs, the gap between software and hardware implementations is already considerable \cite{8114708}.\par
With the advancement in the terminal or edge devices, few contributions are observed in detecting anomalies at the edge or terminal devices. Work in \cite{8170817} present a federated learning approach in which autoencoders are deployed on edge devices to identify anomalies. Utilizing a centralized server as a back-end processing system, the local models are updated and redistributed to the edge devices. However, having a costly back-end server is not a feasible idea for residential surveillance.\par
In \cite{nikouei2018smart}, a lightweight-Convolutional Neural Network (L-CNN) architecture is presented specialized for human detection. The computational cost of the CNN itself is reduced by employing depth-wise separable convolution \cite{howard2017mobilenets} which splits each convolution layer into two parts. Such computational pathways are more suitable for edge devices without much sacrifice on the accuracy of the whole network.
To reduce the memory and storage requirements of using raw image/video data for training, \cite{9116533} used structured, and tensorized time-series features to train a LSTM-based Spatio-temporal model instead of directly working on the raw video frames or non-structured features. Tensor decomposition and an 8-bit trained quantization were performed to achieve deep compression. Experiments on large-scale video-labeled dataset UCFCrime \cite{8578776} is performed using GTX-1080Ti GPU. However, the development and performance analysis of anomaly detector algorithms on CPU-only edge/terminal devices is underexplored.

\section{Methodology}
This section presents the hardware experimental setup and proposed framework for anomaly detection in residential video surveillance.  
\subsection{Hardware Setup}
Our system is intended to detect anomalies in the residential video surveillance streams. The surveillance camera deployment is critical to ensure that the people are within the view and should not be far away to compromise the object detector.\\

\begin{figure}[!t]
\includegraphics[width=0.9\linewidth]{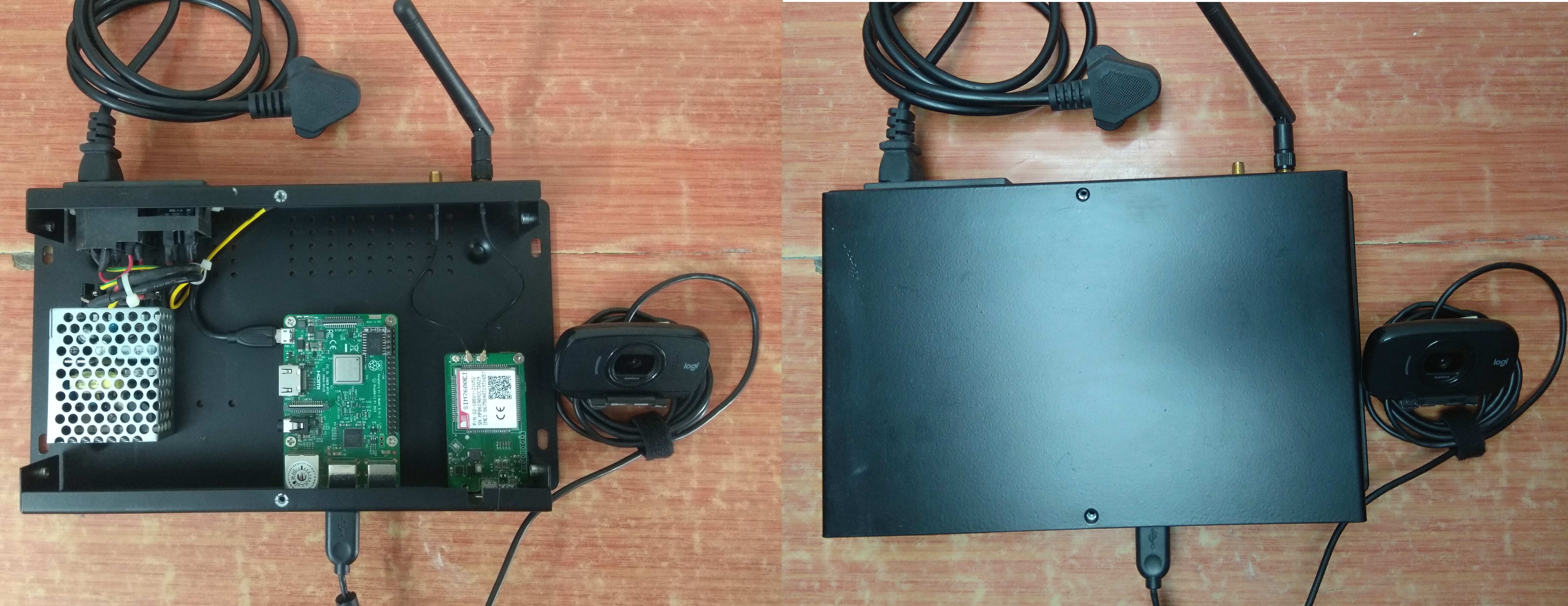}
\centering
\caption{Hardware setup of the proposed system}
\end{figure}

\begin{figure*}[t]
\includegraphics[width=0.825\linewidth]{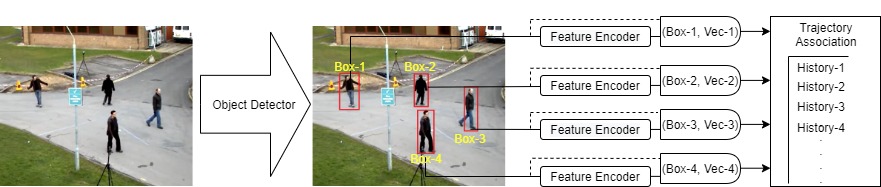}
\centering
\caption{Workflow of the proposed anomaly detector}
\end{figure*}

Our CPU-only edge node is Raspberry Pi 4B, it is a Broadcom BCM2711 SOC with a quad-core-A72 based on ARM v8 architecture. This 64-bit processor runs at 1.5GHz and has a RAM of 4GB. The edge node is actively cooled using an armor aluminum alloy case with a dual cooling fan enclosure; powered and controlled through a GPIO interface. 12,000 mAh Lithium-Ion battery (optional if the power supply is available) is used to power Raspberry Pi and 433MHz LoRa Module. The LoRa module serves as an IoT link between the edge device and the LoRa-based IoT network. We used a fixed view camera, Logitech B525 for recording in full color. An optional, 7-inch LCD is interfaced via HDMI for troubleshooting and visual output. The complete system is packaged in a metal casing for ease of handling and deployment, shown in Figure 2. We use OpenCV, TensorFlow with TensorFlow Lite engine controlled by Python running under Raspbian, a Debian-based operating system optimized for raspberry pi hardware. 

\subsection{Proposed Framework}
The proposed methodology explores spatial association and dynamics between people to represent their behaviour and detect anomalies video surveillance scene. This is possible by utilizing the trajectories of the people in Muti-Object Tracking (MOT) scenario, as it can directly provide object-level interpretations in the scene.
However, posture deformations and heavy occlusions limit the ability of the multi-object tracker. We cope with this limitation by associating trajectories of different objects in the scene. The spatio-temporal feature representation corresponding to each person is developed and used for trajectory association and re-identification. If a person is new and does not have a history in tracking, a fresh identity is assigned and a person that fails to be found over a certain number of frames is considered to have left the scene. Based on the estimated track vectors and the dynamics in them, different anomalies are identified. The workflow of the proposed anomaly detector is shown in Figure 3.

\subsection{Object Detection}
The object detection model is a version of a pre-trained SSD MobileNet v1 \cite{howard2017mobilenets} model optimized for low computational overload. Depthwise separable convolution \cite{sifre2014rigid} technique reduces the computational overload in the MobileNet. A standard convolution is factorized into a depthwise convolution and pointwise convolution which is a 1x1 convolution. In standard convolution, inputs are filtered and combined into one step to generate the output whereas, depthwise separable convolution performs the same operation into two steps, a separate step for filtering and a separate step for combining. This reduces the computations of detphwise separable convolution as compared to the standard convolution by factor of \(\frac{1}{Oc} + \frac{1}{D_\zeta^2}\). Where, \(Oc\) is the number of output channels and \(D_\zeta\) x \(D_\zeta\) is the kernel size. 
We use images of size 300x300 as an input to the MobilNet and from the output, the detection corresponding to the category of 'person' is used for feature extraction and trajectory association.

\subsection{An Association Problem}
The trajectory association must be performed separately on every detected person, which scales the algorithm linearly with the number of persons to be tracked. We use a single hypothesis tracking and data association in consecutive frames for trajectory association. In this framework, the trajectory is defined on the eight dimensional state space (\(u, v, \gamma, h, \dot{x}, \dot{y}, \dot{\gamma}, \dot{h}\)). Where, (\(u,v\)) represents the object bounding box, \(\gamma\) indicates aspect ratio, \(h\) indicates height and (\(\dot{x}, \dot{y}, \dot{\gamma}, \dot{h}\)) indicate the respective velocities in the image coordinates. \par

We use a conventional Kalman filter using liner observation model, and constant velocity motion with object state observation defined by (\(u, v, \gamma, h\)). The trajectory \(T_i^k\) over \(i\)-frames corresponding to a \(k\)-th bounding box detection \(d_{k}\) is considered to be active if \(i<I_{max}\) since last successful trajectory association \(a_k\). Here, \(I_{max}\) is the predefined value and indicate the maximum number of frames without association for which the trajectory is alive. For this, we increment the count \(i\) for each track during kalman filter prediction since the last successful \(a_k\) and reset to 0 at the successful \(a_k\). 

To solve the association problem, we integrate spatial and temporal information by defining two metrics; first, \(C^{'}(l, m)\) representing Mohalanobis distance between the predicted kalman states and newly arrived measurement. Second, \(C^{''}(l, m)\) represents a minimum cosine distance between the appearance descriptors \(r^i_k\in \mathbb{R}^{128}\) computed over each bounding box detection \(d_k\) and set of descriptors computed at each frame for each trajectory \(k\). The combined metric for trajectory association is given by; 
\begin{equation}
    C_{(l,m)}^k = \lambda  c_{k}^{'}(l,m) + (1-\lambda)    c_{k}^{''}(l,m)
\end{equation}
Where, \(\lambda\) is the hyperprameter to control the influence of each metric.  

The motion information \(c_{k}^{'}(l,m)\) is further expanded using Mohalanobis distance as, 
\begin{equation}
    c_{k}^{'}(l,m) = (d_{m}-y_{l})^T S_{l}^{-1} (d_{m}-y_{l})
\end{equation}
Where, we represent the projection of the \(l\)-th track distribution into measurement space (\(y_{l}, S_{l}\)) and \(m\)-th detection \(d_{m} \in\mathbb{R}^4\).
Practically, it calculates the deviation from the mean track location in terms of the standard deviations. The Mohalanobis distance provides accurate association when the motion uncertainty is low. While, the unaccounted camera motion and occlusions induce error in the metric. Further, the predicted state distribution from the kalman filter dose not estimate accurate object location. Therefore, the second metric \(c_{k}^{''}(l,m)\) is integrated in (1). It represent the a minimum cosine distance calculated between object descriptor \(r^i_k\in \mathbb{R}^{128}\) corresponding to each bounding box detection \(d_{k}\) and a set of descriptors \(R_{k} = \{r_{i}^{k}\}_{i=1}^{L_{k}}\) for each track \(k\). So, for \(l\)-th track and \(m\)-th bounding box detection, the metric is defined as, 
\begin{equation}
      c_{k}^{''}(l,m) = min\{1-r_{m}^{T} r_{i}^{l} | r_{i}^{l}\in R_{k} \} 
\end{equation}

In practice, we trained a Convolutional Neural Network (CNN) model as shown in Table-1 on a separate dataset to extract an object descriptor for a given bounding box detection. 
The combined metric (1), serves both aspects of the assignment problem; the Mahalanobis distance investigates object locations based on motion that are especially important for short-term predictions. While, the cosine distance examines appearance information that is useful in recovery of identities after long-term occlusions, especially when motion is less discriminative.
Thus, the associated trajectories are robust to appearance changes, fast motion and occlusions. These trajectories and dynamics in them are employed to detect anomalies in the scene.

\begin{table}[!t]
    \centering
    \caption{CNN Architecture to extract feature descriptor.}
    \label{table_5}
    \begin{tabular}{lccc}
    \hline\noalign{\smallskip}
        Layer &  Size & Stride & Output\\
        \hline\noalign{\smallskip}
        Conv & 3X3 & 1 & 32X64X32\\
        Conv & 3X3 & 1 & 32X64X32\\
        Max Pool & 3X3 & 2 & 32X32X16\\
        Residual & 3X3 & 1 & 32X32X16\\
        Residual & 3X3 & 1 & 32X32X16\\
        Residual & 3X3 & 2 & 64X16X8\\
        Residual & 3X3 & 1 & 64X16X8\\
        Residual & 3X3 & 2 & 128X8X4\\
        Residual & 3X3 & 1 & 128X8X4\\
        Dense  &   &  & 128\\
        BathNormalization & & & 128\\
        \noalign{\smallskip}\hline
    \end{tabular}
\end{table}

\subsection{Anomalies in the residential video surveillance}
Premeditated anomalies are defined based on the person's movement in the typical residential video surveillance streams as follows.\\
• A person standing still for a long time\\
We consider, standing still for a the considerable time does not fit the normal behavior of a human being and it looks suspicious. If the person's tracking trajectory is confined in a very small area for a long time, we define that as an anomaly.\\  
• A person with a fast motion\\
In residential campuses, running is not a usual practice. if a person exceeds the finite fluctuation around the average velocity, we consider this behavior as anomalous.\\
• A person with the circular/spiral movements\\
The spiral motion of a person is very rare and surely suspicious. Generally, such motion is observed when anybody is inspecting something or someone. The tracking trajectory corresponding to a person is utilized, if it is circular in nature the motion is considered as circular/spiral.\\ 
• A person with jumping from the fence or gate\\
Unauthorized entry or trespassing into the residential campus is always attempted by jumping a fence or gate. If the vertical fluctuations in the trajectory exceed the average vertical fluctuations, the situation is declared as an anomaly.\\ 
• Gathering of four or more persons simultaneously\\  
Gathering in a residential campuses is common, however, the simultaneous gathering of four or more persons is not expected as common event and thus considered abnormal. Trajectories with different identities approaching towards the common meet point are identified and declared as anomalies.\\   
• Dispersion of four or more persons simultaneously\\
Dispersion is the same as gathering except for the directions of trajectories are opposite and are extending from the common meeting point.
\subsection{Anomaly Detection}
Consider, the set of predefined anomalies \(A =\{A_{n}\}_{n=0}^{N-1}\) and the estimated set of trajectories \(T =\{T^{k}\}_{k=0}^{K-1}\) corresponding to \(k\) detections. The optimal similarity between them is defined by the function \(w^* :\{0... N-1\} \rightarrow \{0... K-1\}\) that maximizes the similarities between their elements:
\begin{equation}
\begin{split}
    \Theta(w^*) & = \max_{w\in W} \Theta(w)\\
                & = \max_{w\in W} \sum_{j} sim(A(j)T(w(j)))
\end{split}
\end{equation}
Where, \(w(j)\) provides position of the elements in \(T\). In practice, we used a positive cosine distance to calculate similarity in (4). The set \(W\) of possible shifts \(w*\) is updated recursively so that, their computations are efficient and tolerant to some deformations. 
So, if any of the aforementioned situations occur, the anomaly is declared and the alert is sent to the authorized personnel via the LoRa-based IoT-framework. Further, the only data transmitted using LoRa network is the alert of anomaly detection securing complete privacy which is sensitive in residential surveillance. 

\section{Experiment}
\label{sec:4}
The proposed anomaly detection approach employs SSD MobileNet for person detection, as it provides good performance and accuracy. We compared the performance of two versions v1 and v2 of MobileNet at normal CPU clock and also with the overclocked settings on Raspberry Pi 4B. During the tests, the CPU is overclocked from 1500MHz to 1900MHz to analyze the boost in performance of MobileNet at these clock rates as shown in Fig. 4.\par 
We analyzed the performance of feature encoder and trajectory association together on Raspberry pi at normal CPU clock and also at overclocked CPU. We found that the operation took 109ms for a single detection. As the feature encoding and trajectory association are scaled with the number of persons detected in the surveillance scene, the current settings of the operation are not suitable for real-time processing of the surveillance scene with multiple people. Most of the processing time out of 109ms is consumed in the feature encoder which is a CNN trained on MARS dataset \cite{zheng2016mars} with an image input size of 128x64. We retrained it with an image input size of 64x32 and got a speed improvement of about 62\%. With this, the feature encoding and trajectory association took 45ms and even improved by overclocking to 42ms per detection. This setting is considerably faster and practical for computer vision applications running on CPU-only edge devices, as it provides sufficient speed for anomaly detection in residential video surveillance.\par
Setting hardware for the optimum performance,i.e. CPU clock at 1900MHz and feature encoder with input image size 64x32, the overall execution time can be estimated as;\\
(a) Object Detection (OD) in the surveillance frame takes 92ms.\\
(b) Feature Encoding (FE) per detected object takes 38ms.\\
(c) Trajectory Association (TA) takes 4ms.\\ 

\begin{figure}[!t]
\includegraphics[width=0.9\linewidth]{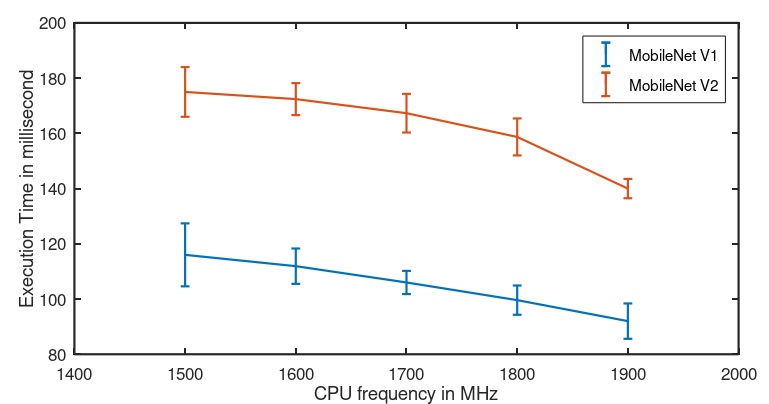}
\centering
\caption{Effect of overclocking on MobileNet performance.}
\end{figure}

Thus, the overall execution time \(\tau\) considering \(d_{k}\) bounding box detections in the surveillance scene is given as,
\begin{equation}
    \tau = 96 + 38*d_{k}
\end{equation}
With this setting, we can detect anomalies with eight detections in the surveillance scene maintaining 2.5 FPS, five detections with 3.5 FPS, two detections with 6 FPS, and one detections with 8 FPS.\\  

\begin{table}[!t]
    \centering
    \caption{Execution with Movidius Neural Compute Stick}
    \label{table_1}
    \begin{tabular}{cccc}
    \hline\noalign{\smallskip}
        Element & Without MNCS & With MNCS & \% Improvement\\
        \noalign{\smallskip}\hline\noalign{\smallskip}
        OD & 92 ms & 73 ms & 20.6\%\\
        FE & 38 ms & 31 ms & 18.4\%\\
        TA & 04 ms & 03 ms & 25.0\%\\
        \noalign{\smallskip}\hline
        
    \end{tabular}
    
\end{table}

Hardware accelerators like, Movidius Neural Compute Stick (MNCS) are very useful to achieve higher processing speed for real-time applications. We used Movidius Neural Compute Stick V1 which is powered by Myriad Vision Processing Unit (VPU) and an AI-optimized chip for accelerating vision computing based on Convolutional Neural Networks (CNN). Integration of MNCS with Raspberry pi 4B provides considerable improvements in the results as shown in Table 2. 
The hardware accelerator (MNCS) boosts the overall performance by about 20\% and facilitates detection of anomalies with a sufficient frame rate for residential surveillance applications.

\section{Result and Discussion}
\label{sec:5}
The experimental setup accepts live camera feed as input to the algorithm. However, to test the algorithm using standard datasets, we set up a system to accept pre-recorded video sequences instead of the live camera feed. We used UCSD Ped1 \cite{UCSD_2010} and UMN \cite{mehran2009abnormal} dataset to test a proposed system both qualitatively and quantitatively.\par
In general, the regularity score whose threshold is manually specified, defines whether the input frame is normal or abnormal. The optimal value of this parameter is very important as higher threshold increases false negative rate, while a lower threshold increases false positive. Thus,  we use the Area Under
Curve (AUC) which is a more suitable metric \cite{luo2017revisit}, \cite{mahadevan2010anomaly}. It measures the performance by changing different thresholds. We adopt frame-level comparison that predicts which frames contain
anomalous events. This is compared to the frame-level ground-truth anomaly annotations to determine the number of true- and false-positive frames.   
Following same protocol, the proposed system is tested extensively and the obtained AUC for the set of experiment is given in Table-3.\par

\begin{table}[!t]
    \centering
    \caption{Anomaly detection scores at different parameter configuration}
    \label{table_2}
    \begin{tabular}{c}
        UCSD Ped-1 Dataset\\ 
        \begin{tabular}{cccc}
        \hline\noalign{\smallskip}
            Feature & max-cos & nms & AUC \\
            Encoder & -distance & -overlap & Score\\
            \noalign{\smallskip}\hline\noalign{\smallskip}
            128x64 &    0.6 &   0.3 &   0.892\\
            128x64 &	0.6 &	0.6 &	0.921\\
            128x64 &	0.6 &	0.9	&   0.873\\
            128x64 &	0.9 &	0.3 &	0.925\\
            128x64 &	0.9 &	0.6 &	0.848\\
            128x64 &	0.9 &	0.9 &	0.802\\
            64x32 &	0.6 &	0.3 &	0.885\\
            64x32 &	0.6 &	0.6 &	0.913\\
            64x32 &	0.6 &	0.9 &	0.838\\
            64x32 &	0.9 &	0.3 &	0.918\\
            64x32 &	0.9 &	0.6 &	0.851\\
            64x32 &	0.9 &	0.9 &	0.835\\
            \noalign{\smallskip}\hline
        \end{tabular}\\
    \end{tabular}\\
    \vspace{0.4cm}
    \begin{tabular}{c}
        UMN Dataset\\ 
           \begin{tabular}{cccc}
           \hline\noalign{\smallskip}
            Feature & max-cos & nms & AUC \\
            Encoder & -distance & -overlap & Score\\
            \noalign{\smallskip}\hline\noalign{\smallskip}
            128x64 &	0.6 &	0.3 &	0.923\\
            128x64 &	0.6 &	0.6 &	0.895\\
            128x64 &	0.6 &	0.9 &	0.931\\
            128x64 &	0.9 &	0.3 &	0.927\\
            128x64 &	0.9 &	0.6 &	0.891\\
            128x64 &	0.9 &	0.9 &	0.886\\
            64x32 &	0.6 &	0.3 &	0.931\\
            64x32 &	0.6 &	0.6 &	0.917\\
            64x32 &	0.6 &	0.9 &	0.891\\
            64x32 &	0.9 &	0.3 &	0.936\\
            64x32 &	0.9 &	0.6 &	0.872\\
            64x32 &	0.9 &	0.9 &	0.882\\
            \noalign{\smallskip}\hline
           \end{tabular}\\
    \end{tabular}
    
\end{table}

We tested several parameter configurations; (a) the ‘non-maximum suppression’ (nms-overlap) threshold for object detector to avoids spurious overlapping detection boxes, (b) for feature encoder, we tested size of feature encoder 64x32, 128x64, and 256x128 and (c) the max-cos-distance threshold for trajectory association.\par 
It is observed that, reducing the size of input feature encoder dose not affect anomaly score significantly but reduces the processing time drastically. We also observed, lower non-maximum-suppression effectively suppress some spurious boxes that are generated by clusters of people and achieves slightly improved performance.\par 
Further, for trajectory associations, setting the max-cos-distance threshold to higher value achieves better performance. Lower values of the threshold may result into unwanted identity swaps as the cosine distances between two encoded feature vectors are not large enough to differentiate people wearing similar color clothing.\par

\begin{table}[!t]
    \centering
    \caption{Comparative results of anomaly detection on CPU-only edge device}
    \label{table_3}
    \resizebox{8.5cm}{!}{%
    \begin{tabular}{lcc}
    \hline\noalign{\smallskip}
        Methods & UCSD Ped-1 Dataset & UMN Dataset\\
        \noalign{\smallskip}\hline\noalign{\smallskip}
        \cite{8422970} & 0.781 & 0.816\\
        \cite{xue2020anomaly} & 0.863 & 0.872\\
        \cite{nawaratne2019spatiotemporal}  & 0.890 & 0.912\\
        \cite{luo2017remembering} & 0.772 & 0.785\\ 
        \textbf{Proposed} & \textbf{0.918} & \textbf{0.936}\\
        \noalign{\smallskip}\hline
       
    \end{tabular}
     }
\end{table}

Based on the results obtained for different parameter configurations, we set the best configuration for the proposed system as feature encoder size to 64x32, max-cos-distance to 0.9, and nms-overlap threshold to 0.3. With this optimum configuration, we analyzed the power consumption for running and sleeping states at different CPU clock rates as shown in Fig. 5. We also noticed that the core temperature spikes up to about 68$^\circ$C without active cooling and about 59$^\circ$C with active cooling. The effect of overclocking on CPU cores is shown in Fig. 6. However, without active cooling, further increase in the CPU clock, increases the core temperature and halts the system due to overheating.

Finally, with the optimum setting, we compare the proposed approach with existing methods on UCSD Ped-1 and UMN dataset as shown in Table-4. Images capturing anomalies in the surveillance scenes are shown in Fig. 7. Results demonstrate that the proposed method outperforms the other approaches by achieving higher scores. The spatio-temporal feature encoding effectively works in associating trajectories and ensures consistency in person identities even in complex scenes, deformed postures, and partial occlusions. Furthermore, the improvement in time efficiency, though not suitable for large-scale surveillance environments, is considerable for facilitating anomaly detection using CPU-only edge devices in residential video surveillance systems.

\begin{figure}[!t]
\includegraphics[width=8cm, height=4cm]{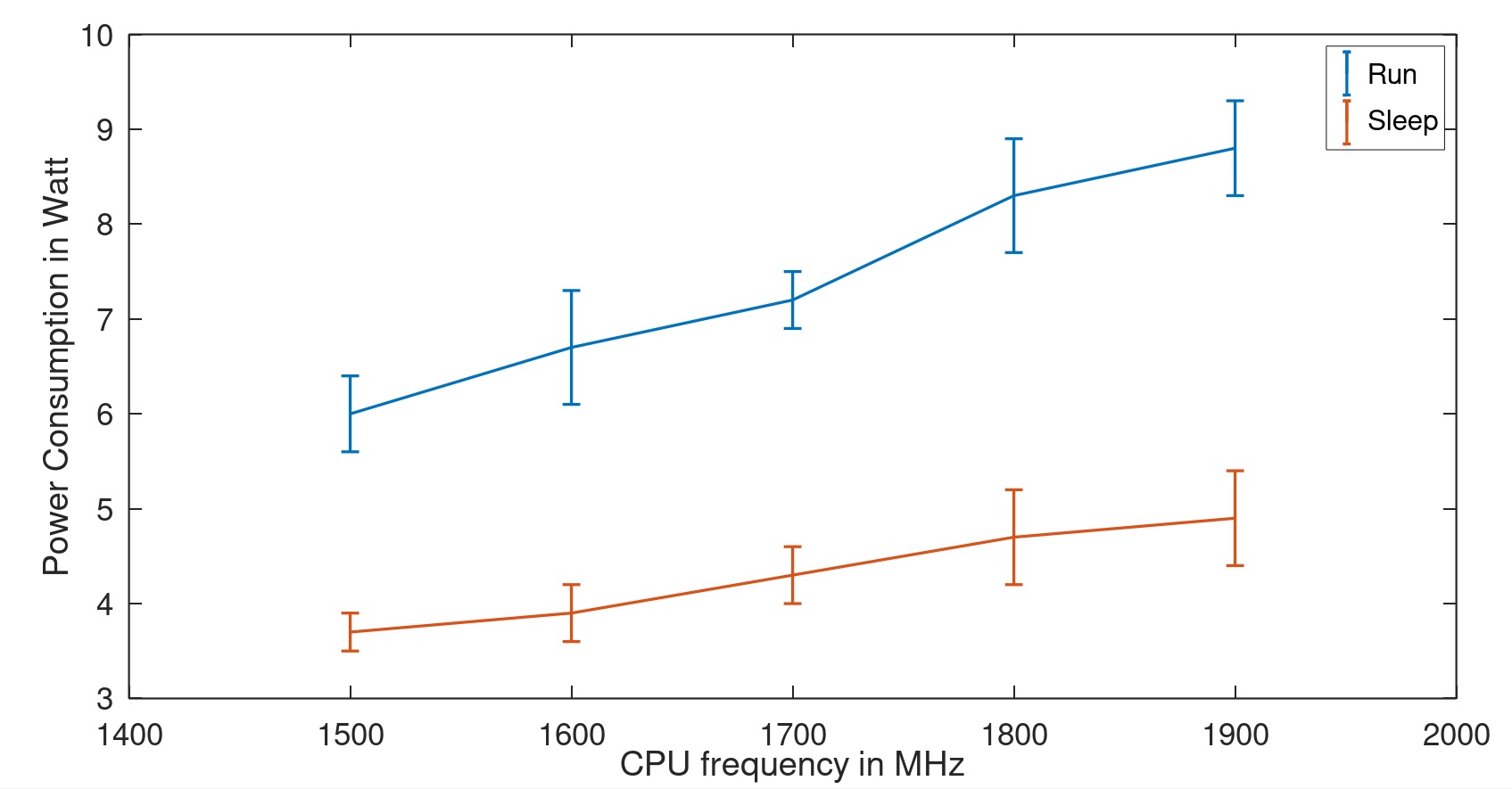}
\centering
\caption{Power consumption in running and sleeping state.}
\end{figure}

\begin{figure}[!t]
\includegraphics[width=8cm, height=4cm]{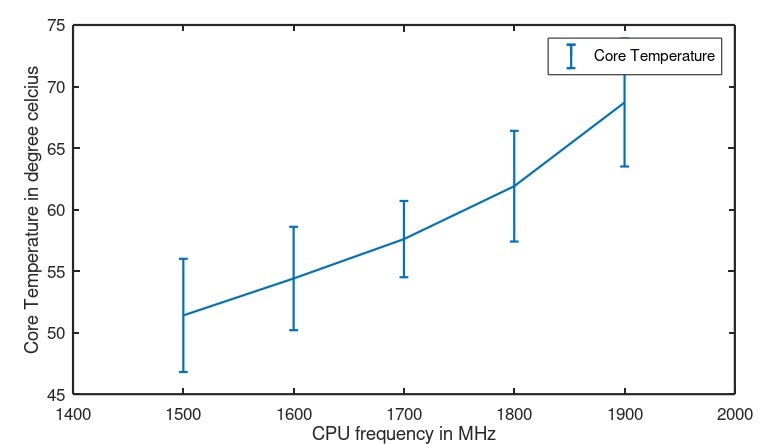}
\centering
\caption{Effect of overclocking on core temperatures.}
\end{figure}

\begin{figure}[!t]
\includegraphics[width=0.9\linewidth]{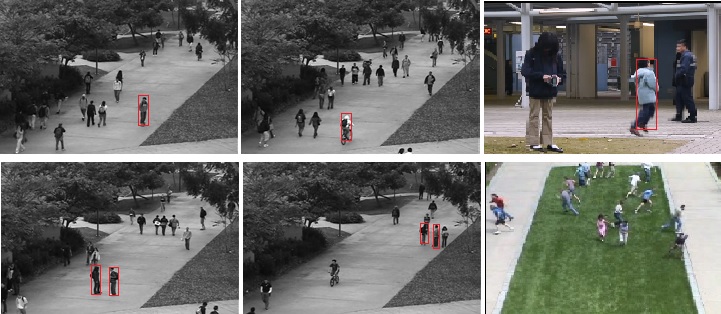}
\centering
\caption{Anomalies in surveillance videos from UCSD Ped-1 and UMN datasets}
\end{figure}
\section{Conclusion}
\label{sec:6}
In this work, to make the intelligent surveillance system highly portable and cost effective, we used CPU-only edge devices to detect anomalies in residential video surveillance. The associated trajectories and dynamics in them are utilized to detect abnormal behavior of people in surveillance scenes. The trajectory association corresponding to each detection is performed by integrating spatial and temporal information. The spatio-temporal integration is governed by two metrices that provide promising results in cluttered environment. The alerts for the anomalies are sent through LoRa network maintaining complete privacy in the residential environment. Through several experiments, we optimized the feature encoder to achieve optimized processing time and FPS by slightly compromising on the performance. The experimental results on datasets validate the competitive advantage of the proposed approach on CPU-only edge devices. Moreover, we also establish that the use of hardware accelerators can significantly increase the FPS in detecting anomalies on CPU-only edge devices.

\ifCLASSOPTIONcaptionsoff
  \newpage
\fi

\bibliography{JAIHC.bib}
\bibliographystyle{IEEEtran}
\vspace{-1cm}

\end{document}